\documentclass[letterpaper, 10pt, conference]{ieeeconf}
\IEEEoverridecommandlockouts
\usepackage{graphics} 
\usepackage{amsmath} 
\usepackage{amssymb}  
\usepackage{siunitx}
\usepackage{bm}
\usepackage{url}

\usepackage{booktabs} 
\usepackage{makecell}
\usepackage{placeins}
\usepackage{todonotes}
\presetkeys{todonotes}{inline,backgroundcolor=white}{}

\usepackage{xspace}
\usepackage{color}
\makeatletter
\DeclareRobustCommand\onedot{\futurelet\@let@token\@onedot}
\def\@onedot{\ifx\@let@token.\else.\null\fi\xspace}

\def\eg{\emph{e.g}\onedot} 
\def\ie{\emph{i.e}\onedot}

\makeatother

\DeclareMathOperator*{\argmin}{arg\,min}
\newcommand{\norm}[1]{\left\lVert#1\right\rVert}
\usepackage{dsfont}
\newcommand{\mathbbm}[1]{\mathds{#1}}

\newcommand{\mypara}[1]{\vspace{1mm}\noindent\textbf{#1}}

\definecolor{forestgreen}{rgb}{0.13, 0.55, 0.13}
\definecolor{frenchblue}{rgb}{0.0, 0.45, 0.73}
\newcommand{\blueobj}{\textcolor{frenchblue}{blue object}}
\newcommand{\greenobj}{\textcolor{forestgreen}{green object}}

\title{\LARGE \bf
Learning Object Manipulation Skills from Video via Approximate Differentiable Physics
}

\author{Vladim\'ir Petr\'ik$^{\clubsuit}$ \and %
Mohammad Nomaan Qureshi$^{\diamondsuit}$ \and %
Josef Sivic$^{\clubsuit}$ \and %
Makarand Tapaswi$^{\diamondsuit}$ %
\thanks{$^{\clubsuit}$ CIIRC, Czech Technical University in Prague {\tt\footnotesize \{vladimir.petrik, josef.sivic\}@cvut.cz}}%
\thanks{$^{\diamondsuit}$ CVIT, IIIT Hyderabad, India {\tt\footnotesize \{mohammad.nomaan@research., makarand.tapaswi@\}iiit.ac.in}}%
\thanks{This work was supported by the European Regional Development Fund under the project IMPACT (reg. no. CZ.02.1.01/0.0/0.0/15\_003/0000468).} %
}

\begin{document}

\maketitle
\thispagestyle{plain}
\pagestyle{plain}

\begin{abstract}

We aim to teach robots to perform simple object manipulation tasks by watching a single video demonstration.
Towards this goal, we propose an optimization approach that outputs a coarse and temporally evolving 3D scene to mimic the action demonstrated in the input video.
Similar to previous work, a differentiable renderer ensures perceptual fidelity between the 3D scene and the 2D video.
Our key novelty lies in the inclusion of a differentiable approach to solve a set of Ordinary Differential Equations (ODEs) that allows us to approximately model laws of physics such as gravity, friction, and hand-object or object-object interactions.
This not only enables us to dramatically improve the quality of estimated hand and object states, but also produces physically admissible trajectories that can be directly translated to a robot without the need for costly reinforcement learning.
We evaluate our approach on a 3D reconstruction task that consists of 54 video demonstrations sourced from 9 actions such as \textit{pull something from right to left} or \textit{put something in front of something}.
Our approach improves over previous state-of-the-art by almost 30\%, demonstrating superior quality on especially challenging actions involving physical interactions of two objects such as \emph{put something onto something}.
Finally, we showcase the learned skills on a Franka Emika Panda robot.

\end{abstract}

\section{Introduction}

Learning from Demonstrations (LfD) uses manual demonstrations in a target domain to teach a robot new skills~\cite{argall2009survey}.
The learning signal in LfD is different from classic reinforcement learning (RL) that often involves an engineer designing and fine-tuning a separate reward function for each robot skill or task such that the reward guides the learning process towards the desired solution~\cite{wu2017doom,james2020rlbench}.

In LfD, the demonstrations are usually obtained by robot teleoperation~\cite{zhang2018imitationComplexTasks} or direct manual guiding~\cite{bazzi2020manualguidance}.
While this may be easier than tweaking a reward function, depending on the environment, capturing demonstrations at scale can be a cumbersome process.
Drawing inspiration from instructional videos on Youtube~\cite{miech2019howto100m} or educational videos on learning platforms that help humans learn and acquire new skills at scale\footnote{\eg, Skillshare~\url{https://www.skillshare.com/}}, recent works have proposed methods to leverage video demonstrations to teach robots simple object manipulation skills~\cite{sermanet2018tcn,shao2020concept2robot,petrik2020real2sim}.
Contrary to robot guiding or teleoperation that requires special infrastructure, learning from videos relies on human demonstrations recorded at homes or other environments of daily activity.


\begin{figure}[t]
\centering
\includegraphics[width=\linewidth]{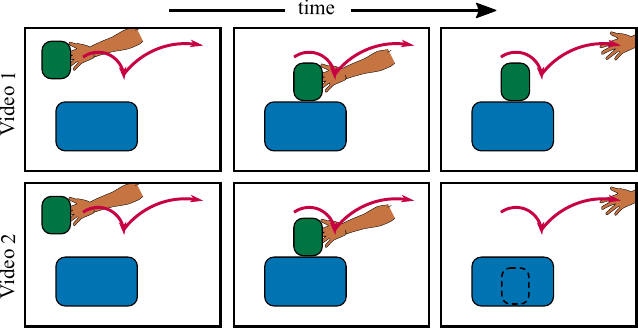}
\caption{
\textbf{By watching the first two frames, can you guess whether the person puts the \greenobj{} \emph{on top of} or \emph{behind} the \blueobj{}?}
We illustrate three frames of two video demonstrations, in which a front-view camera looks at a hand (in orange) and two objects: a static object placed on the ground (in blue) and a manipulated object (in green).
The red curve visualizes the motion of the hand as time flows from left to right.
In both the videos, the hand moves the \greenobj{} from free space (first frame) to what appears like \emph{above} the \blueobj{} in a 2D camera projection (middle frame).
However, without seeing the last frame, it is impossible to tell if the \greenobj{} is behind the \blueobj{} or if they are at the same distance.
Once the hand releases the object (third frame), we can estimate their relative positions by using basic physics properties such as \emph{gravity} and \emph{contact interactions}.
Note that modeling these physics properties allows distinguishing between the two cases as the \greenobj{} can no longer float in the air.
For example, in video~1, we see that the \greenobj{} stays above the \blueobj{} indicating that it depicts the action \emph{on top of}, while in video~2, the \greenobj{} \emph{falls} and is occluded by the \blueobj{} (shown as a dashed line) indicating the \emph{put behind} action.
The depth ambiguity can thus be reduced by incorporating basic laws of physics.
}
\label{fig:motivation}
\end{figure}

A recent work, \emph{Real2Sim}~\cite{petrik2020real2sim}, shows that it is possible to learn simple object manipulation skills from a single video using a two stage approach.
A first stage reconstructs coarse 3D hand and object state trajectories using a differentiable renderer that produces object and hand silhouettes to closely resemble the video segmentation masks; while the second stage uses RL to learn a robot policy to mimic these trajectories.
However, they also introduce multiple hard-coded priors to lift the 2D videos to 3D reconstructed scenes such as assumptions about the object being attached to the ground.
We build upon~\cite{petrik2020real2sim} and remove the need for such hard-coded prior knowledge by introducing a physics-based regularization that also results in improved performance for reconstructing trajectories.
In fact, we hypothesize that modeling physics is not only convenient, but also necessary to reduce ambiguities in perception.
We motivate our hypothesis through Fig.~\ref{fig:motivation} that illustrates an example of visual ambiguity in \emph{putting something \underline{behind} or \underline{on top of} something}.

The goal of our work is to teach robots simple object manipulation skills from a single video.
To do this effectively, we propose to incorporate differentiable physics into the optimization process.
We show that modeling physical laws such as gravity and inter-object interactions (including hand/gripper and object) reduces the depth ambiguity problem leading to faithful reconstructions.
Similar to~\cite{petrik2020real2sim}, we use simple geometric shapes to represent the hand and the objects as a coarse approximation of the video scene and show that such a representation is sufficient to reconstruct actions with a high accuracy.
We evaluate our approach on a benchmark consisting of 9 actions and focus on hand and object state trajectory reconstruction from a single video.
Incorporating the proposed physics-based regularization increases model performance by almost 30\% and also provides physically admissible hand/gripper motions that can be directly translated to a robot environment, mitigating the need for RL.

\mypara{Contributions.}
We summarize our contributions as follows.
(i) We propose a differentiable, physics based regularization, that is combined with motion optimization to learn robotic skills from videos;
(ii) Our approach is suitable for learning from a single video demonstration and in fact produces physically admissible trajectories that afford re-targeting to a robot environment without costly reinforcement learning; and
(iii) Our approach results in a dramatic performance improvement over the baseline, especially on actions involving physical interactions between two-objects such as \emph{put something onto something}.
We will make the code and data publicly available\footnote{\url{https://github.com/petrikvladimir/video_skills_learning_with_aprox_physics}}.

\section{Related Work}
We briefly review recent work in modeling hand-object interactions, differentiable physics, and learning from video demonstrations.

\mypara{Modeling of hand-object interactions.}
Accurate 3D modeling of hand-object interactions has gained significant attention owing to the development of MANO~\cite{romero2017mano}, an articulated deformable model of the hand, whose shape and pose parameters can be estimated reliably at a high frame-rate~\cite{zhou2020monocular}.
This model is used to train a robust, image-based estimator of hand-object interactions~\cite{shan2020understanding} that not only predicts bounding boxes of the hand and the object but also estimates their contact status.
We use these contact estimates to optimize the hand/gripper grasping signal.

Given an input image/video, estimation of the hand pose together with a 3D mesh for an unknown object is explored in~\cite{hasson2019handobject}. 
An object-hand penetration loss is introduced as a physics-based regularization.
Other approaches in the same direction jointly optimize the hand's shape and pose and a known object's pose and scale~\cite{cao2021handobjectwild,qin2021dexmv}.
Collision and interaction losses are used here for regularization.
Note that the above works attempt to estimate accurate hand poses from an image/video, which remains challenging especially for low quality images or occluded hands or objects.

As our goal is to teach robots simple manipulation skills, we focus on the temporal evolution of the hand and objects and are able to model them with coarse geometric primitives.
We also replace the modeling of accurate hand-object interactions (\eg~finger contacts~\cite{hasson2019handobject}) with an approximate grasp signal without requiring to actually model hand grasps.


\mypara{Differentiable physics.}
A differentiable physics model enables computation of gradients for some physical simulation (forward process).
This gradient can be used for optimization of latent physical properties such as mass or friction coefficients~\cite{jatavallabhula2021gradsim, kandukuri2020learning,asenov2019vid2param}.
Various works have explored differentiable physics in context of multi-body systems~\cite{physics1}, articulated bodies~\cite{physics2}, or soft-multi body systems~\cite{physics4}. 
Differentiable physics has also been applied to learn manipulation skills such as cutting~\cite{cutting} or throwing~\cite{throwing}.
In the proposed work, we encode basic laws of physics through equations of motion represented by Ordinary Differential Equations (ODEs).
We use a neural ODE solver~\cite{chen2018neuralode} that allows us to compute the gradient of the forward simulation similar to recent differentiable physics engines~\cite{jatavallabhula2021gradsim}.
In addition, we also introduce a control signal based on whether the hand is grasping an object as an event that modifies the equations of motion for the object.

\mypara{Learning from demonstrations.}
Using demonstrations to guide optimization in  RL~\cite{argall2009survey,hazara2016contact,vecerik2017demonstrations} is a viable alternative to cumbersome and task-specific reward shaping.
The demonstrations are usually obtained by teleoperation or manual guiding~\cite{zhang2018imitationComplexTasks,bazzi2020manualguidance} and consist of robot motions performed in the target environment.
Learning from video mitigates the need for demonstrations in the target environment and instead leverages video representations to guide the learning through
reward estimation (\ie~inverse reinforcement learning) methods~\cite{sermanet2017learnfromvideo,sermanet2017tcn,chen2021learning,das2021model},
human to robot domain translation~\cite{liu2018imitationobservation,smith2020avid,xiong2021learning}, or
a task classifier used as a reward function~\cite{shao2020concept2robot}.
Different from the above works, our optimization approach uses differentiable physics to produce physically admissible trajectories that can be re-targeted to a robot without the need for RL.

Perhaps closest to our work is the learning from videos approach proposed in~\cite{petrik2020real2sim} that estimates a coarse temporally evolving state from the video and uses it as a reward signal for RL.
While our work is similar in spirit, we are able to circumvent some of the hard-coded prior knowledge in~\cite{petrik2020real2sim} by modeling simple physics laws like gravity and contact interactions.
This not only leads to physically admissible trajectories that mitigate the need for RL, but also results in a large improvement in the quality of estimated trajectories.

\section{Learning from Video via Approximate Physics}
\begin{figure*}[t]
\centering
\includegraphics[width=0.8\linewidth]{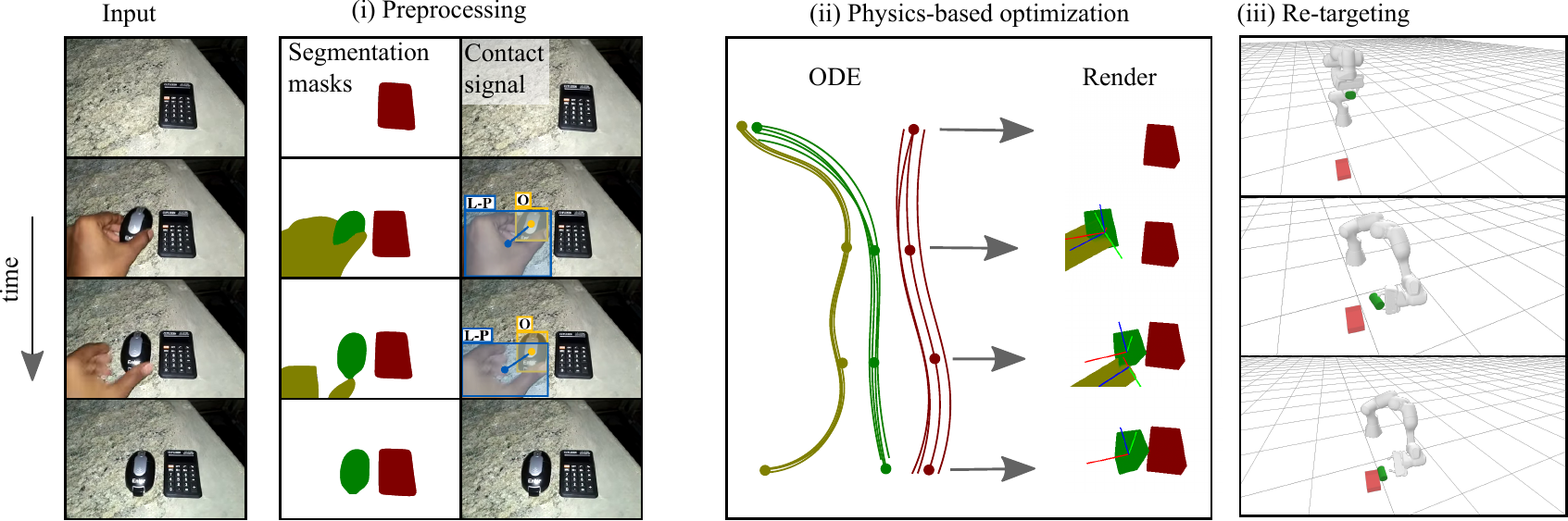}
\vspace{-2mm}
\caption{
Our proposed pipeline is illustrated for four keyframes of a video demonstration corresponding to the action \emph{put something next to something}.
The pipeline consists of:
(i) a \emph{video preprocessing} step which extracts segmentation masks and information about the hand-object contact;
(ii) a \emph{physics-based optimization} step that solves physics equations of motion (ODEs), renders the states via a differentiable renderer and computes and minimizes losses based on perceptual similarity to the video demonstration. In the figure panel, the curves represent trajectories and dots are the discrete timesteps at which loss is computed; and
(iii) a \emph{re-targeting} step which computes the robot trajectory from optimized Cartesian trajectories.
}
\vspace{-3mm}
\label{fig:pipeline}
\end{figure*}

The goal of our work is to teach robots simple object manipulation skills from a single video demonstration.
We split the learning approach into three steps: 
(i) Video preprocessing extracts hand and object segmentation masks and estimates if the hand is in contact with the object;
(ii) Optimization leads to estimation of physically admissible hand and object state trajectories that perceptually resemble the video demonstration; and
(iii) Re-targeting maps the estimated motion onto the robot.
The whole pipeline with intermediate results is visualized in Fig.~\ref{fig:pipeline}.

\subsection{Notation and problem formulation}
\mypara{Video demonstration.}
The video consists of $N$ frames that are denoted by~$I_k, k \in \{1, \ldots, N \}$.
The subscript~$k$ denotes discrete time-steps corresponding to the frames of the video.
The video demonstration shows a person (observed as a hand due to an ego-centric view) performing some action consisting of a simple object manipulation.

\mypara{Simulated scene.}
We find approximate but physically consistent state trajectories for the hand and objects that resemble the action demonstrated in the input video.
To this end, we approximate the hand with a fixed sized cylinder of length $\SI{40}{\centi\meter}$ and radius $\SI{4}{\centi\meter}$.
The shape and dimensions are chosen to approximate an adult human hand.
The hand position and orientation are denoted by symbols~$\bm x^h_t \in \mathbb{R}^3$ and~$R^h_t \in SO(3)$, for $t \in [0, T]$, where $T$ is duration of the video demonstration.
Different from the discrete time-steps $k$, the symbol $t$ is used for continuous time and we operate in the continuous time range $[0, T]$ for the optimization.

The objects in the scene are approximated by cuboids and their sizes are estimated by the optimization.
To model the actions chosen from our data, we assume that there is one manipulated object in the scene and optionally one immovable or static object.
However, note that the proposed approach can be extended to an arbitrary number of static objects.
The manipulated object's position, orientation, and size are denoted by $\bm x^o_t \in \mathbb{R}^3$, $R^o_t \in SO(3)$, and $\bm s^o\in \mathbb{R}^3$ respectively.
Note that size $\bm s^o$ is independent of time.

The optional immovable object is not directly acted upon in the demonstration video and in fact can be thought of as being part of the scene layout.
However, the manipulated object may collide with it.
We refer to this object as the collision-aware object and use superscript $c$.
Its position, orientation, and size are denoted by $\bm x^c \in \mathbb{R}^3$, $R^c \in SO(3)$, and $\bm s^c\in \mathbb{R}^3$ and are all independent of time.

The complete state of the scene at any time $t$ is denoted
by~$\bm z_t = \left( \bm x^h_t, \, R^h_t, \, \bm x^o_t, \, R^o_t, \, \bm s^o, \, \bm x^c, \, R^c, \, \bm s^c \right)$,
\ie~it consists of the hand pose at time $t$, the manipulated object's pose at time $t$ and its size, and the static object's pose and size.

\mypara{Problem formulation.}
Given camera parameters $\bm c$, the state of the scene can be rendered into an image via differentiable rendering~\cite{ravi2020pytorch3d,liu2019softras}.
We denote the rendering process as a function~$f_\text{render}(\bm z_t, \bm c)$, where $\bm c$ describes both the intrinsic and extrinsic camera parameters.
As mentioned earlier, our reconstructed scene is an approximation of the video and the rendered images are not directly comparable to the RGB pixels of the video demonstration.
Instead, we extract binary masks for the hand~$M^h_k$, the manipulated object~$M^o_k$, and the static object~$M^c_k$ from the input video.
Similar to~\cite{petrik2020real2sim}, we define the perceptual distance at the level of rendered silhouettes and segmentation masks.

We formulate our goal as an optimization problem to find a joint solution for all time-evolving states of the scene and the static camera parameters such that the scene rendering resembles the video demonstration while respecting some simple laws of physics:
\begin{align}
\bm z_t^\ast, \bm c^\ast = \argmin\limits_{\bm z_t\in\mathcal{Z}, \bm c} \mathcal{L}(\bm z_t, f_\text{render}(\bm z_k, \bm c), M^h_k, M^o_k, M^c, \tau_k) \, , \label{eq:formulation}
\end{align}
where 
$\bm z_t^\ast$ and $\bm c^\ast$ represent the optimal solution of time-evolving states of the scene and camera parameters,
$\bm z_k$ represents the state at a discrete time-step corresponding to the video frames,
$\tau_k$ is the hand-object contact signal estimated from the video and used to guide the optimization,
$\mathcal{L}$ is the loss function to minimize,
and $\mathcal{Z}$ represents the set of all physically admissible trajectories.
The details of the loss function are discussed in subsequent sub-sections.
The main difficulty in optimizing Eq.~\eqref{eq:formulation} is generation of physically admissible trajectories $\bm z_t$ -- we solve this locally by adopting advances in neural ordinary differential equation~(ODE) solvers~\cite{chen2018neuralode}.

\subsection{Differentiable physics}
Prior to describing the equations of motion that are used as part of the ODE solvers, we briefly touch upon the physics approximations.

\mypara{Approximate physics.}
The approximation adopted by our proposed method is threefold:
(i)~We use coarse geometric models for the hand and the objects which allows us to compute the signed distance function in a differentiable manner.
(ii)~We use an elastic collision model which approximates the real-world contacts and results in a small penetration of the objects in contact; and
(iii)~we approximate real-world friction with a tangential velocity reduction that depends linearly on the object velocity. The real-world friction cone is therefore not modeled.

\mypara{Hand ODE.}
We consider the hand cylinder as a control entity that can use one of its ends (tip) to grasp and release the object.
The motion of the hand/gripper is driven by an active entity and is therefore not directly governed by the same physical constraints as that of objects (for example, the gravity does not affect hand).
We approximate the continuous motion of the hand and represent the time derivatives of the hand position, orientation, and grasp signal using a spline that acts as a regularizer:
\begin{equation}
\bm v^h_t, \, \bm \omega^h_t, \, \dot g^h_t  = f_\text{spline}(t) \, ,
\end{equation}
where
$\bm v^h_t$ is the linear velocity of the hand,
$\bm \omega^h_t$ is the spatial angular velocity,
and $g^h_t \in \mathbb{R}$ is the grasp signal that represents the hand's ability to hold the object.
If the grasp signal is positive and the hand is in contact with the object (\ie~the hand tip is inside the object), the object is grasped by the hand and can be released only when the grasp signal becomes negative.
The differential equations describing the hand motion are:
\begin{align}
    \bm{\dot x}_t^h = \bm v_t^h, \quad \dot R_t^h = (\bm \omega^h_t)^\wedge R^h_t \, , \label{eq:eom_hand}
\end{align}
where the hat operator $^\wedge$ is used to construct a skew-symmetric matrix from a given vector~\cite{murray2017mathematical} (see Eq. 2.4) such that $\bm a \times \bm b = (\bm a)^\wedge \bm b$.

\mypara{Object ODE.}
Contrary to the hand, the state trajectory of the manipulated object from its starting position is determined either by the hand or by laws of physics such as gravity and collision interactions.
For brevity we will omit $t$.

The object's equations of motion are captured as 4 components.
The linear velocity $\dot{\bm x}^o$ is influenced by the hand's linear and angular velocities when grasped.
The linear acceleration $\dot{\bm v}^o$ consists primarily of interaction impulses due to collisions $\bm j_v$, or acceleration due to gravity $\bm g$ when not grasped.
Similarly, the angular velocity $\dot{R}^o$ sums up contributions of the individual hand and object velocities when grasped and the angular acceleration $\dot{\bm \omega}^o$ models impulses due to collisions $\bm j_\omega$:
\begin{align}
\dot{\bm x}^o &=
      \begin{cases}
        \bm v^h + \bm \omega^h \times (\bm x^o - \bm x^h) + \bm v^o, & \text{if object grasped, } \\
        \bm v^o , & \text{otherwise, } \label{eq:eom_obj_xdot}
      \end{cases} \\
\dot{\bm v}^o &=
      \begin{cases}
        -k_{j,v} \bm v^o + \bm j_{v}, & \text{if object grasped, } \\
        -\bm g + \bm j_{v}, & \text{otherwise, } \label{eq:eom_obj_vdot}
      \end{cases} \\
\dot R^o &=
      \begin{cases}
        \left( \bm \omega^h + \bm \omega^o\right)^\wedge R^o, & \text{if object grasped, } \\ 
        (\bm \omega^o)^\wedge R^o, & \text{otherwise, } \label{eq:eom_obj}
      \end{cases} \\
\dot{\bm \omega}^o &=
      \begin{cases}
        -k_{j,\omega} \bm \omega^o + \bm j_\omega, & \text{if object grasped, } \\
        \bm j_\omega, & \text{otherwise, } \label{eq:eom_obj_omegadot}
      \end{cases} 
\end{align}
where $k_{j,v}$ and $k_{j,\omega}$ are positive constants that control how quickly the collision impulse decays for grasped objects.

\mypara{Object impulses.}
We first compute the impulse experienced at each of the $P$ sampled points on the manipulated object's boundary when it collides with some object (or ground) $\mathcal{O}$.
Let $\bm x_i$ and~$\bm v_i$ be the position and velocity of a point on the boundary,
and $\bm n$ be the normal at a point on the surface of $\mathcal{O}$ closest to $\bm x_i$.
The magnitude of the impulse $m_i$ at point $i$ is proportional to the penetration distance of a point $\bm x_i$ inside the collision object $\mathcal{O}$
and to the colliding velocity, \ie~the velocity that would result in collision if integrated over time.
The impulse at each point on the boundary $\bm j_i$ combines the surface normal vector with an approximation of friction through a tangential velocity that is present only for non-zero magnitudes:
\begin{align}
m_i &= \max \left\{-k_p d(\bm x_i, \mathcal{O}) - k_v \min\left\{\bm v_i \cdot \bm n,\, 0\right\} , 0 \right\} \, , \label{eq:eom_impulse_mag}\\
\bm j_i &=  m_i \bm n - \mathbbm{1}(m_i) \mu \bm v^\text{tangential}_i \, ,
\end{align}
where~$k_p$ and~$k_v$ are positive scalar constants controlling the stiffness of the collision, and
$\mu$ is a non-negative scalar that acts like a coefficient of friction, and
$\mathbbm{1}(m_i)$ is one for non-zero $m_i$ and zero otherwise.
The term~$\bm v_i \cdot \bm n$ represents the velocity in the contact's normal direction, which is clamped by the $\min$ operator to generate impulses only for colliding velocities.


The linear and angular impulses are combined over the individual sampled points as:
\begin{align}
\bm j_v &= \sum\limits_{i=1}^{P} \bm j_i, \quad \bm j_w = \sum\limits_{i=1}^{P}\bm q_i \times \bm j_i \, ,
\end{align}
where $\bm q_i$ is a vector pointing from the center of the object to the point $\bm x_i$.
In our setup, recall that the manipulated object may only collide with either the ground plane $z = 0$ or an optional immovable object.

\mypara{Discrete events.}
The object's equations of motion depend on a discrete variable that indicates if the object is grasped by the hand or not (see Eqs.~(\ref{eq:eom_obj_xdot}-\ref{eq:eom_obj_omegadot})).
We define an \emph{event} function and capture this discrete variable through a change in sign as follows:
\begin{align}
f_{\text{event}} = \max\{ d_{\text{hand-object}}, \, -g^h \}\, , \label{eq:eom_events}
\end{align}
where $d_{\text{hand-object}}$ represents the signed distance function between the hand tip and the manipulated object's boundary and $g^h$ is the estimated hand grasp signal.
The object is grasped if and only if the hand grasp signal is positive and the signed distance is negative, indicating that the hand tip is \emph{inside} (in contact with) the object.
The recent advances in differentiable ODE solvers~\cite{chen2021eventfn} allow us to automatically detect the zero-crossings of $f_{\text{event}}$ and update the discrete object grasp status accordingly in a differentiable way.

\mypara{Integrating the equations of motion.}
Equations~(\ref{eq:eom_hand}-\ref{eq:eom_events}) describe the ODEs that can be integrated over time given the initial state of the manipulated and static objects and a known hand velocity specified by the controlling spline.
The physically consistent state is therefore computed based on the equations of motion and this state can be used as a constraint in the optimization objective of Eq.~\eqref{eq:formulation}.

\subsection{Optimization}
As stated in Eq.~\eqref{eq:formulation} we optimize the loss over state trajectories and camera parameters.
The trajectory is computed by integrating the ODE based on the start state and hand motion that are being optimized.
However, not all parameters are well conditioned in our tasks.
For example, the rotation about the hand axis can take arbitrary values (when it is not in touch with the manipulated object) owing to the rotation symmetry of the cylinder that represents the hand.
We constrain some of the ill-conditioned parameters by a regularization loss or by removing corresponding degrees of freedom from the optimized parameters as discussed next.

\mypara{Optimization state space.}
For all rotation matrices, we use a 6D representation~\cite{zhou2019rot6d}.
We optimize the following \emph{hand parameters}: velocities $(\bm v^h, \bm \omega^h, \dot g^h) \in \mathbb{R}^7$ at the control points of the spline, and the initial state of the hand $(\bm x^h_0, R^h_0, g^h_0) \in \mathbb{R}^{10}$.
Thus, the hand is parameterized by $7 N_\text{spline} + 10$ degrees of freedom (DOF), where $N_\text{spline}$ represents the number of spline control points.

The \emph{manipulated object parameters} include the object start position, orientation and size (12~DOF).
The initial object velocity is set to zero.
We assume that the \emph{immovable object} (cuboid) lies on the ground on one of its faces.
Thus, it is parameterized by a 2D position, rotation about the gravity axis and its size (6~DOF).

The \emph{camera intrinsic parameters} are determined by a standard perspective camera~\cite{ravi2020pytorch3d}.
The effect of intrinsic focal length is not observable as it can be compensated by simultaneously changing the position of the hand.
As we use a fixed size cylinder to represent the hand, for a fixed focal length, we can estimate the pose of the hand.
We observe that the inaccuracy in estimating the camera intrinsic parameters is not critical as it would either affect the distance of the hand from the camera origin (influence of focal length) or result in small perceptual distortions that are negligible as compared to our coarse cuboid and cylinder approximations (due to camera distortions).
To estimate the \emph{camera's extrinsic parameters}, we model the camera height and the elevation angle and assume that the gravity vector is perpendicular to the rows of the captured image, which is often true for egocentric demonstrations.
All other camera pose DOF are compensated by offsetting the position of the scene elements.

\mypara{Loss.}
The loss function from Eq.~\eqref{eq:formulation} is divided into three parts:
(i) The \emph{perceptual loss} ensures that the segmentation mask $M_k$ at frame $k$ resembles the rendered silhouette produced by the differentiable renderer $f_\text{render}(\bm z_k)$:
\begin{align}
\mathcal L_\text{per} =\sum\limits_{k=1}^N 
\sum\limits_{j\in\{h,o,c\} } 
w^j_\text{per} \mathcal J (M^j_k, f_\text{render}(\bm z_k)) \, , \label{eq:loss_per}
\end{align}
where $j$ is used to sum over the hand, the manipulated object, and the collision object with weights $w^h_\text{per}, w^o_\text{per}, w^c_\text{per}$ respectively; and
$\mathcal J$ represents the distance-Intersection-over-Union loss (dIoU) adopted for segmentation masks from work on bounding box regression~\cite{zheng2020diou}.
The dIoU has a low score when the two masks align well.

(ii) The regularization loss encourages a small hand and object velocity and prevents needless motion. It also specifies that the object should not move when it is not visible, \ie~when $\mathbbm{1}(M^o_k)$ is zero:
\begin{align}
\mathcal L_\text{reg} = \sum\limits_{k=1}^N 
 w^{o}_\text{reg} (1-\mathbbm{1}(M^o_k)) \| \bm v^o_k + \bm \omega^o_k \|^2 +
 w^{h}_\text{reg} \| \bm v^h_k + \bm \omega^h_k \|^2
\, , \label{eq:loss_reg}
\end{align}
where $w^{o}_\text{reg}, w^{h}_\text{reg}$ weight the norm of object and hand velocities.

(iii) Finally, the contact loss uses the hand-object grasping status $\tau_k$ estimated from the demonstration to guide the grasping signal $g^h_k$.
Note that the loss is not applicable when $\tau_k = 0$.
\begin{align}
\mathcal L_\text{con} = w^d_{\text{con}} \sum\limits_{k=1}^N
\mathbbm{1}(\tau_k) \norm{\max\{d_\text{hand-object}, 0 \}}^2 +  \nonumber \\
\mathbbm{1}(\tau_k) \norm{\min\{ g^h_k, 0 \}}^2 
+(1-\mathbbm{1}(\tau_k)) \norm{\max\{ g^h_k, 0 \}}^2,
\end{align}
where $w^d_{\text{con}}$ weights the contact loss.
The first term penalizes a positive signed distance function $d_\text{hand-object}$ while the second term penalizes a negative hand grasp signal for frames where the hand is detected to be in contact (\ie~$\mathbbm{1}(\tau_k)=1$).
The last term penalize positive grasp signal for frames not in contact.
In our experiments, we found that the contact loss was necessary to guide the hand towards actual grasping.
Without it, the optimizer did not discover that grasping the object leads to a lower perceptual loss as the local gradient does not provide sufficient information.

Finally, the combined loss $\mathcal L = \mathcal L_\text{per} + \mathcal L_\text{reg} + \mathcal L_\text{con}$ is minimized using the Adam optimizer~\cite{kingma2015adam}.

\mypara{Initialization.}
We observe that the optimization of the Eq.~\eqref{eq:formulation} from a random initial state is quite challenging, perhaps due to the presence of multiple local minima -- several learned trajectories ignored the object and only learned to move the hand.
Therefore, a decent initialization is necessary to start the optimization.

We estimate the initial parameters by only optimizing the perceptual loss and the distance term of the contact loss in the camera frame of reference without any physics.
Then, we use optimization via physics to find an appropriate camera pose and to further fine-tune the initialized trajectories.
We show via experiments that the initialization alone is insufficient for finding physically consistent state trajectories, however, it does provide a reasonable starting point for the optimizer.

\subsection{Re-targeting to the robot arm}
The physics-based optimization has an advantage as it can provide physically admissible trajectories of the hand, object, and the contact state.
For transferring the hand Cartesian motion to the robot we optimize the trajectory of the robot's joints such that the robotic gripper follows the estimated hand trajectory.
The optimization takes into account the constraints given by the robot and the environment, \eg~the joint limits or the robot link collisions.
We use contact change to enforce the robotic gripper to grasp the object at one of the pre-defined local grasp-poses or to release the object at a specific time.

\section{Experiments}

We start this section with a brief overview of the video demonstrations and the metrics that are used to evaluate our approach.
We then provide some implementation details, followed by a quantitative evaluation.

\begin{table*}[t!]
\centering
\tabcolsep=1.5mm
\caption{
Quantitative comparison of the estimated trajectories.
}
\vspace{-2mm}
\begin{tabular}{lccccccccccc}
\toprule
\makecell{Method} 
& \makecell{Pull left\\ to right}
& \makecell{Pull right\\ to left}
& \makecell{Push left\\ to right}
& \makecell{Push right\\ to left }
& \makecell{Pick \\ up}
& \makecell{Put \\ behind}
& \makecell{Put in \\ front of}
& \makecell{Put \\ next to}
& \makecell{Put \\ onto}
& \textbf{Total}
& \makecell{\textbf{Success} \\ \textbf{rate}} \\
\midrule
Real2Sim~\cite{petrik2020real2sim} orig. masks & 5/6 & 4/6 & 5/6 & 3/6 & 1/6 & 3/6 & 1/6 & 0/6 & 0/6 & 22/54 & \SI{40}{\percent} \\
Real2Sim~\cite{petrik2020real2sim} STCN masks & 6/6 & 6/6 & 6/6 & 6/6 & 3/6 & 5/6 & 0/6 & 0/6 & 1/6 & 33/54 & \SI{61}{\percent} \\
Proposed initialization & 3/6 & 4/6 & 2/6 & 2/6 & 0/6 & 5/6 & 5/6 & 5/6 & 0/6 & 26/54 & \SI{48}{\percent}  \\
Proposed approach & 6/6 & 6/6 & 6/6 & 6/6 & 4/6 & 6/6 & 6/6 & 6/6 & 3/6 & 49/54 & \SI{90}{\percent}  \\
\bottomrule
\end{tabular}
\vspace{-5mm}
\label{tab:quantitative}
\end{table*}

\subsection{Dataset and Metric}
\vspace{-1mm}
\mypara{Something Something dataset.}
Following~\cite{petrik2020real2sim}, we evaluate the hand and object state trajectories on 9 actions and a total of 54 video demonstrations (6 per action) from the \emph{Something Something} dataset~\cite{goyal2017something}.
An example of each action can be seen in Fig.~\ref{fig:results}.
Please refer to the supplementary video for additional qualitative results.

\mypara{Evaluation metric.}
We define a metric that captures various properties of the hand and object states, such as the amount of displacement for \emph{pull/push} actions or the angle of the hand to disambiguate between \emph{pull} and \emph{push} actions.
For brevity, we will only discuss additional requirements included in this work, but encourage the reader to review Appendix B of~\cite{petrik2020real2sim} for a quick visual overview of the metrics.
One additional criteria applicable to all actions and all timesteps is to verify that the object is not below the ground.

\emph{Pull and push.}
For these single object actions, we require that the object moves in the correct direction for at least~\SI{5}{\centi\meter} with the hand in the proper orientation, \ie~in-front-of or behind the object in the direction of motion respectively.
Our additional constraint applies to the object motion in the gravity axis and requires the object to be lifted less than the size of the object in that same axis.
This verifies that object is not lifted and placed but is actually pushed or pulled instead.
This was not required in~\cite{petrik2020real2sim} due to the hard-coded object height during optimization.

\emph{Pick up.}
We require the object to be lifted above the ground by at least~\SI{1}{\centi\meter} as in~\cite{petrik2020real2sim}.
In addition, to ensure that our proposed model does not cheat, we require that the object is on the table before the action starts.

\emph{Put.}
For the \emph{put next to}, \emph{put in front of}, and \textit{put behind} actions we verify that the manipulated object is placed at the appropriate side of the static object at the end of the motion.
We also require the manipulated object's $z$-dimension at the end of the motion to be below the static object's height in $z$.
This ensures that the object is not held in the air.
For the \emph{put onto} action, we verify that the center of the manipulated object projected to the ground is inside the static object's boundary projected to the ground.
Given a uniform density and convex objects, this ensures that the manipulated object will not topple over the static one.
In addition, at the end of the motion, we check that the manipulated object's bottom face is close to the static object's top face.


\subsection{Implementation details}
\mypara{Video segmentation.} 
Many video segmentation methods require an initial annotated frame which they use to propagate the labels~\cite{oh2019stm,cheng2021modular,hu2018videomatch}.
We annotate polygon bounding boxes for the hand and objects in just \emph{one} frame where all objects are visible.
This happens to be the center frame of the action part of the video (\ie~when the hand is touching the object) for most videos (except a couple cases where the object is occluded).
We use an off-the-shelf pretrained Space-Time Correspondence Network (STCN)~\cite{cheng2021stcn} model, and propagate the labels across the video (forward and backward in time) starting from the central keyframe.
Note that while it is possible to replace the manual segmentation by an automatic one (\eg~using MaskRCNN~\cite{he2017maskrcnn}),
these models do not discover blobs and may miss some of the household objects used in our videos resulting in significant errors in trajectory reconstruction.
To provide a fair comparison against the current state-of-the-art baseline~\cite{petrik2020real2sim}, we re-evaluate~\cite{petrik2020real2sim} on our improved segmentation masks.
As a post-processing step we keep only the largest connected component (blob) and filter out spurious noise in the estimated segmentation masks.

\mypara{Contact estimation.}
We use a hand-object detector model~\cite{shan2020understanding} to estimate whether the hand is in contact with an object.
The method estimates a binary value (in contact or not) for each frame of the video.
Since the hand, or the object, or both, may not be visible from the beginning or at the end of the video, we design a simple but effective multi-step filtering heuristic.
(i) We first smooth the predictions using a median filter.
(ii) If the object is moving at the beginning of the video (based on segmentation masks), we say that the hand-object are in contact until the first negative detection  from~\cite{shan2020understanding}. A similar strategy is used at the end of the video.
(iii) Finally, we check if the release signal towards the end of the video can be brought forward in time as the contact estimation model makes errors when the hand is close to the object. This too is achieved through median filtering.

\subsection{Performance of trajectory estimation}
We evaluate the proposed approach on the same set of 54 videos as used in~\cite{petrik2020real2sim}. 
We use the metric, designed in the previous section to evaluate the trajectories generated by the Real2Sim approach~\cite{petrik2020real2sim} and compare them to the proposed physics-based optimization.
The results are shown in Tab.~\ref{tab:quantitative}.

\mypara{Impact of video segmentation.}
The first two rows of the table evaluate the impact of improving segmentation masks.
Masks obtained through STCN used in this work are cleaner and result in better overall performance compared to the original masks from~\cite{petrik2020real2sim} (40\% vs. 61\%).
Note that this comes at the cost of drawing 2-3 polygon annotations per video demonstration, which we think is negligible as compared to actually performing the demonstration.

\mypara{Impact of initialization.}
We see that the initialization scheme is effective at obtaining rough trajectories (48\%).
However, it is worse than Real2Sim~\cite{petrik2020real2sim} with the same masks (61\%).
This is primarily due to the assumptions and priors encoded in Real2Sim.
Nevertheless, a decent initialization acts as a launchpad for our physics-based optimization approach that achieves a 90\% success rate.
Our approach is able to reconstruct valid trajectories for all videos of 7 actions, and only falters for 5 of 54 videos in two challenging actions \emph{pick up} and \emph{put onto}.

\begin{figure*}[t]
\centering
\includegraphics[width=0.98\linewidth,trim={1cm 0 0 0},clip]{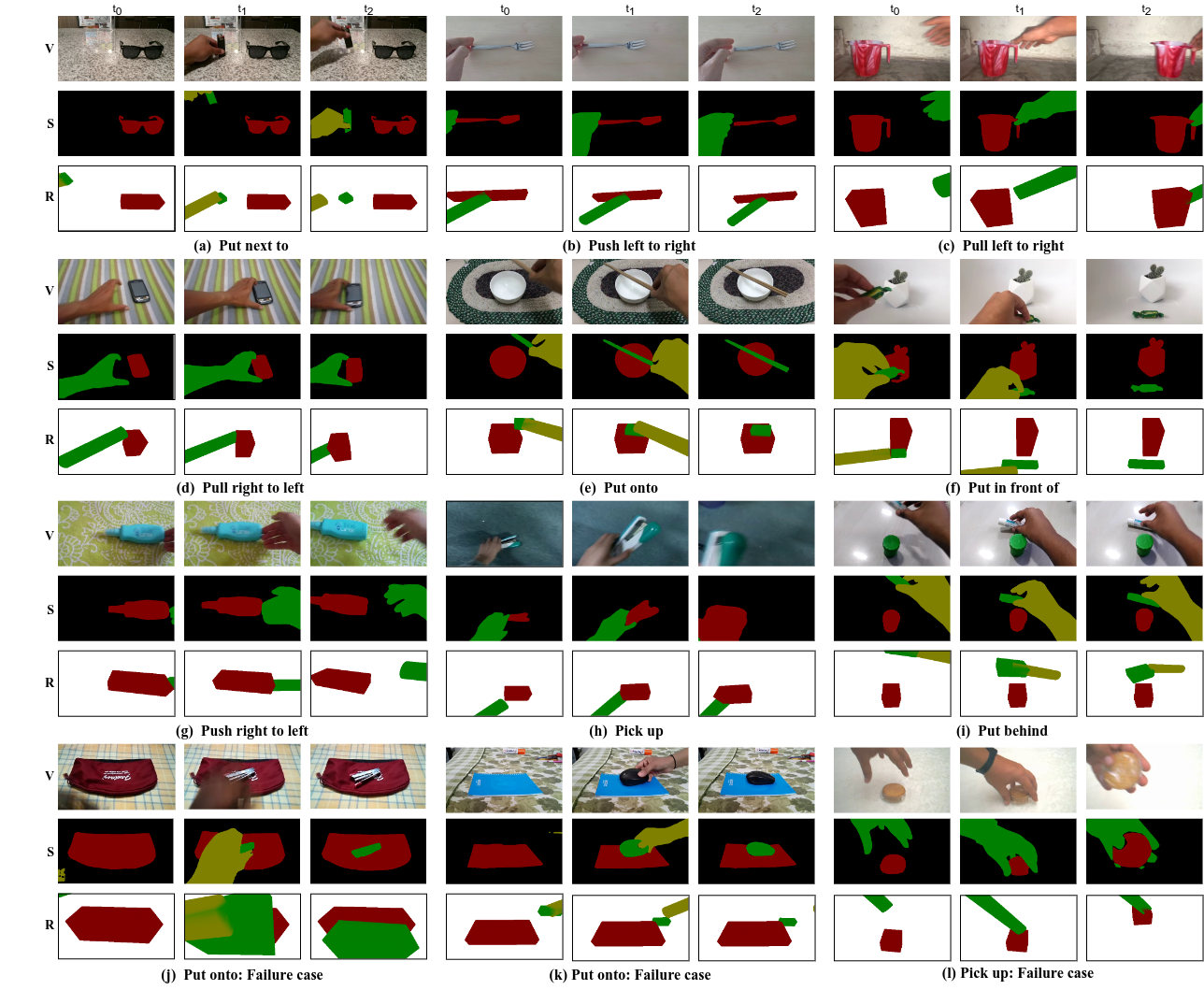}
\vspace{-4mm}
\caption{Qualitative results showing 3 frames of the video demonstration (V), their predicted segmentation masks (S) and the corresponding rendered frame (R) after physics-based optimization.
An example of a successful trajectory completing the target task is shown for all 9 actions from (a) to (i).
In the last row (j) and (k) show failure cases for the \emph{put onto} action, while (l) shows a failed example for the \emph{pick up} action.}
\vspace{-5mm}
\label{fig:results}
\end{figure*}

\mypara{Qualitative analysis.}
Fig.~\ref{fig:results} presents examples of the video frames, segmentation masks, and the rendered output of the 3D scene for a few frames of each demonstration.
The first 9 panels (a)-(i) illustrate successful outputs, one for each of the 9 actions.
The last three panels, (j)-(l) present failure cases for the two actions mentioned above.
We see that failures are primarily due to fast hand motion or errors in segmentation.

\subsection{Re-targeting to the robot arm}
Even with an approximate physics-based regularization, we are able to obtain trajectories that can be directly translated to the robot arm.
Please refer to the supplementary video for additional details of this experiment on our robot.

\section{Conclusion}

We proposed an approximate physics-based optimization approach to reconstruct a temporally evolving 3D scene that mimics a video demonstration.
Our key contribution was to include a differentiable solver for ODEs of hand and object motion that are controlled by a grasp signal modeled as an event.
Together with a differentiable renderer, we presented an approach to recover physically consistent trajectories that were perceptually similar to the input video and can be also directly re-targeted to a robot.
Evaluation on 9 diverse single and two-object actions showed large improvements in the quality of estimated trajectories as compared against previous state-of-the-art.
We successfully transferred these skills to our robot and include examples in the supplementary video.

\mypara{Limitations.}
The main limitation of the proposed method is the difficulty in propagating gradients to optimize the contact/grasp signal making the approach
sensitive to the initialization of the grasp signal.
However, the initialization and refinement strategy presented in the implementation details allow us to achieve high performance.
Other limitations include
requirement for manual annotation of a single-frame,
motion modeling for only one object, and
inability to work with fine actions that may require the full dexterity of the hand due to the coarse cylinder approximation.



\bibliographystyle{IEEEtran}
\bibliography{shortstrings,references}

\addtolength{\textheight}{-0cm}

\end{document}